\documentclass[conference]{IEEEtran}
\IEEEoverridecommandlockouts

\usepackage{cite}
\usepackage{amsmath,amssymb,amsfonts}
\usepackage{algorithm}
\usepackage{algorithmic}
\usepackage{array}
\usepackage{booktabs}
\usepackage{makecell}
\usepackage{multirow}
\usepackage{stfloats}
\usepackage{tabularx}
\usepackage{textcomp}
\usepackage{xcolor}
\usepackage{graphicx}
\usepackage{url}
\usepackage[draft]{hyperref}
\usepackage{verbatim}
\ifCLASSOPTIONcompsoc
\usepackage[caption=false,font=normalsize,labelfont=sf,textfont=sf]{subfig}
\else
\usepackage[caption=false,font=footnotesize]{subfig}
\fi

\def\BibTeX{{\rm B\kern-.05em{\sc i\kern-.025em b}\kern-.08em
    T\kern-.1667em\lower.7ex\hbox{E}\kern-.125emX}}

\makeatletter
\newcommand{\linebreakand}{
  \end{@IEEEauthorhalign}
  \hfill\mbox{}\par
  \mbox{}\hfill\begin{@IEEEauthorhalign}
}
\makeatother

\begin{document}

\title{Enhancing Biomedical Named Entity Recognition via Multiple Programming Languages Instruction Tuning and Ensemble Method\thanks{\textsuperscript{*} Corresponding author}}

\author{
\IEEEauthorblockN{
Songtao Li\textsuperscript{1},
Yijia Zhang\textsuperscript{1,*},
Jianyuan Yuan\textsuperscript{2},
Shidi Zhang\textsuperscript{1},
Fengyu Zhang\textsuperscript{3},
Hongfei Lin\textsuperscript{4}
}

\vspace{0.5em}

\begin{tabular}{c}
\textsuperscript{1}School of Information Science and Technology, Dalian Maritime University, Dalian, China \\

\textsuperscript{2}School of Automation, Beijing Institute of Technology, Beijing, China \\

\textsuperscript{3}School of Computer Science and Engineering, Northeastern University, Shenyang, China \\

\textsuperscript{4}School of Computer Science and Technology, Dalian University of Technology, Dalian, China \\[0.5em]

\{devin\_lee, zhangyijia, zhangshidi\}@dlmu.edu.cn, \\
jianyuany@bit.edu.cn, zhangfengyu@mails.neu.edu.cn, hflin@dlut.edu.cn \\[0.3em]

\end{tabular}
}

\maketitle

\begin{abstract}
Instruction tuning has become a common paradigm for applying large language models (LLMs) to biomedical named entity recognition (BioNER). However, existing instruction-tuning approaches still face two key challenges. First, conventional natural-language instructions typically serialize BioNER annotations as flat textual outputs, providing limited structural constraints for typed entity extraction. Second, high-quality biomedical annotations are limited, and learning from a single serialized output form may restrict structural diversity and reduce model robustness. Although external biomedical knowledge can be introduced to alleviate data scarcity, it often requires costly resource construction.
To address these challenges, we propose MITE, a \textbf{M}ultiple Programming Languages \textbf{I}nstruction \textbf{T}uning and \textbf{E}nsemble method for BioNER. MITE reformulates BioNER as a structure-to-structure generation task by representing both instructions and entity outputs in code-formatted representations. Specifically, each training instance is transformed into multiple programming-language formats, including Python, C++, and Java, while preserving the same underlying entity semantics. These language-specific representations provide structurally diverse supervision without requiring external biomedical knowledge or additional annotations. During inference, MITE aggregates predictions from different code formats through an entity-level voting strategy, reducing language-specific prediction variance and improving robustness. Experiments on three widely used BioNER datasets demonstrate that MITE achieves competitive performance against representative BERT-based and LLM-based baselines, attaining the best results on two benchmarks. Ablation studies further validate the effectiveness of multi-language representations.
\end{abstract}


\section{Introduction}

Biomedical named entity recognition (BioNER) focuses on recognizing biomedical entities such as diseases, chemicals, genes, and proteins from unstructured biomedical text \cite{DBLP:journals/bib/SongLLZ21}.
As a fundamental task in biomedical information extraction, BioNER provides structured entity-level knowledge that supports a wide range of downstream applications, including biomedical relation extraction \cite{zhang2024not} and biomedical question answering \cite{DBLP:conf/acl/Nguyen19}.
Therefore, reliable BioNER is a key foundation for extracting, organizing, and exploiting biomedical knowledge from large-scale scientific literature and clinical text.

\begin{figure}[!t]
\centering
\includegraphics[width=0.99\columnwidth]{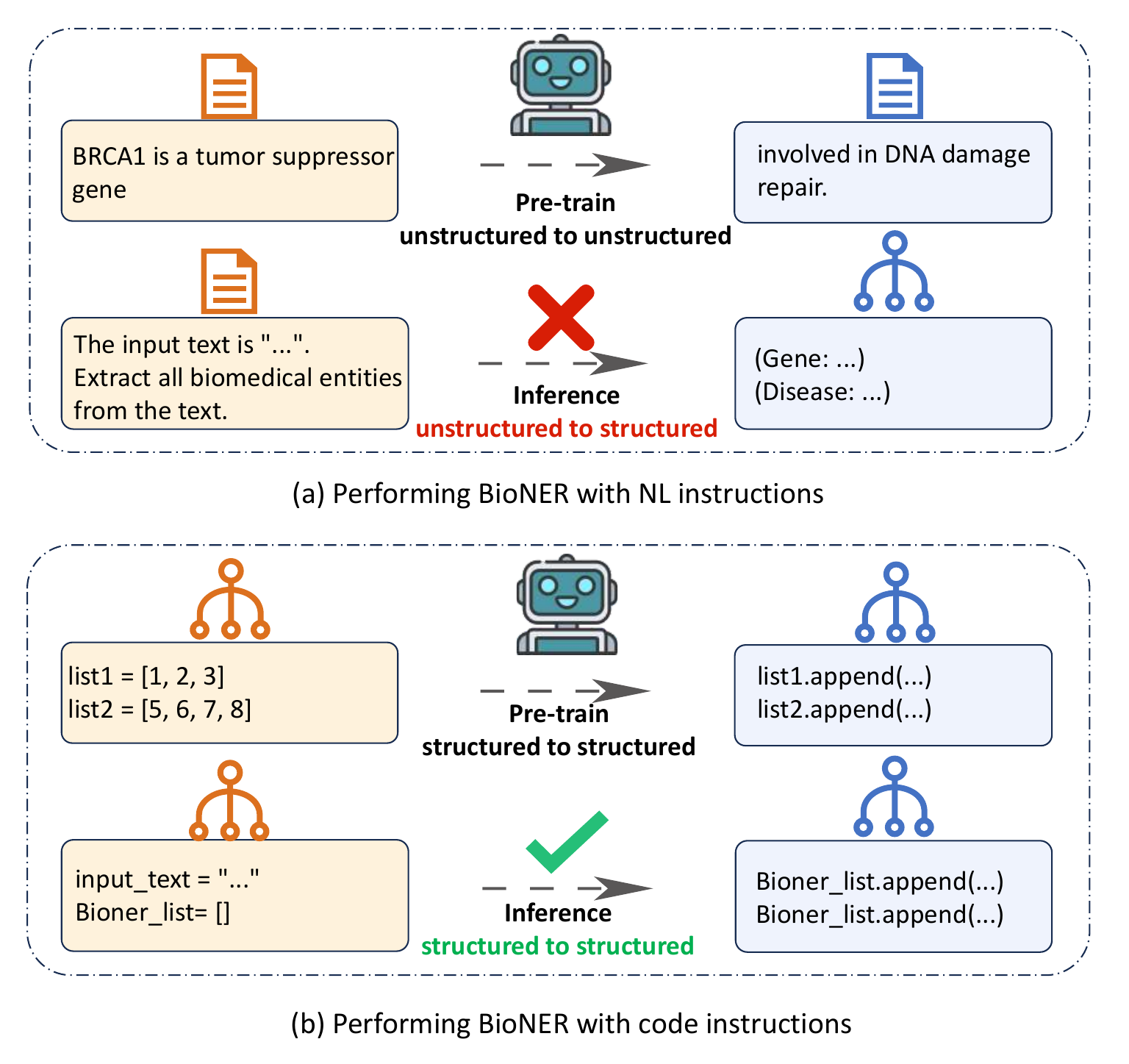}
\caption{Examples of performing structured BioNER task with NL and Code instructions. In contrast to prompting with NL instructions, we utilize structured Code instructions to mitigate the output discrepancy between the pretraining and inference stages.}
\label{fig:NL_vs_code}
\end{figure}

Recent advances in large language models (LLMs) have opened up new opportunities for BioNER by enabling models to follow task-specific instructions and generate flexible textual outputs.
To adapt general-purpose LLMs to biomedical entity extraction, supervised fine-tuning (SFT) has been widely explored, where BioNER instances are converted into instruction-following formats and used to teach models how to extract entities from biomedical text \cite{keloth2024advancing,liu2025resource,calcina2026extracting}.
Under this paradigm, natural language instructions are commonly used to describe the extraction task, specify the target entity types, and guide the model to produce structured predictions.
To further improve extraction performance, several recent studies incorporate external biomedical knowledge or domain-specific resources, allowing models to better capture specialized terminology and biomedical semantics \cite{bian2024vaner,cocchieri2025openbioner}.

Despite notable progress in BioNER, existing studies still suffer from two main limitations.
First, BioNER is inherently a span-type structure prediction task, requiring models to identify entity boundaries and assign semantic types under a predefined schema.
However, current instruction-tuning methods usually formulate entity extraction as natural-language-to-text generation, where entity annotations are serialized into flat textual outputs.
Such a formulation weakens the alignment between typed entity structures and the code-like structured patterns learned by LLMs during pretraining, limiting their ability to exploit explicit structural regularities for BioNER prediction, as illustrated in Fig.~\ref{fig:NL_vs_code}(a).
Second, high-quality biomedical annotations are limited, and existing instruction-tuning methods usually learn BioNER from a single serialized output form. This restricts the structural diversity of training signals and may make models sensitive to specific output formats rather than robust entity semantics. Although external biomedical knowledge can be introduced to alleviate data scarcity, such resources are often costly to construct, limited in coverage, and difficult to maintain.

To address the aforementioned challenges, we propose MITE, a Multiple Programming Languages Instruction Tuning and Ensemble method for BioNER. MITE reformulates BioNER from natural-language-to-text generation into a structure-to-structure generation task, where both task instructions and entity outputs are represented in code-formatted structures. Entity mentions and their semantic types are expressed as explicit code-style objects, providing a constrained output representation that better aligns span-type BioNER prediction with the code-like structured patterns learned by LLMs during pretraining, as illustrated in Fig.~\ref{fig:NL_vs_code}(b).
To further improve robustness, MITE transforms each BioNER instance into multiple programming-language representations, including Python, C++, and Java. These representations introduce structural diversity through different syntax forms, object construction patterns, and list operations, without requiring external biomedical knowledge or additional annotations. During inference, MITE aggregates predictions from different code formats through entity-level voting, reducing language-specific prediction variance and producing more stable BioNER outputs.

In summary, the contributions of this paper are as follows:
\begin{itemize}
\item We reformulate BioNER as a structure-to-structure code generation task, which represents span-type entity predictions as explicit code-style objects and better aligns BioNER with the structured patterns learned by LLMs during pretraining.

\item We transform original BioNER samples into multiple programming-language representations, constructing a structurally diverse training corpus without relying on external biomedical knowledge or additional annotations.

\item We introduce a multi-language entity-level ensemble strategy that aggregates predictions from different code representations to improve the stability and reliability of BioNER outputs.

\end{itemize}

\section{Related Work}

\subsection{Biomedical Named Entity Recognition}

BioNER has progressed from feature-based systems to neural architectures and domain-specific pretrained language models. Early systems relied on handcrafted rules, biomedical dictionaries, and manually designed features, requiring substantial expert knowledge and generalizing poorly to unseen terminology~\cite{savova2010mayo}. Neural models based on CNNs and BiLSTMs subsequently reduced manual feature engineering by learning contextual and character-level representations from annotated corpora~\cite{ren2018neural}. More recently, pretrained models such as BioBERT, BlueBERT, ClinicalBERT, and PubMedBERT have achieved strong performance by learning domain-specific representations from large-scale biomedical corpora~\cite{lee2020biobert,gu2021domain}.

Despite these advances, conventional BioNER methods predominantly follow the sequence-labeling paradigm and require dataset-specific adaptation. This limits their transferability across biomedical corpora with different entity types, annotation guidelines, and label granularities. Multi-task learning methods improve cross-dataset transfer by jointly training on multiple corpora~\cite{caruana1997multitask,khan2020mt}, while AIONER further maps heterogeneous annotations into a unified label space~\cite{luo2023aioner}. However, directly combining multiple datasets may introduce annotation inconsistencies because the same biomedical mention can be assigned different labels or omitted across corpora. Achieving cross-dataset generalization while preserving dataset-specific label semantics therefore remains challenging.

\subsection{LLM-Based Information Extraction}

LLMs have promoted a shift toward generative and instruction-following approaches for specialized biomedical tasks \cite{xu2025echogpt,zhu2026ultrag}. In the information extraction domain, recent studies have explored zero-shot prompting~\cite{averly2025entity, isaradech2025zero}, few-shot demonstrations~\cite{liu2025eicl}, and supervised instruction tuning~\cite{wang2023instructuie, zhou2024universalner} to guide LLMs in extracting structured information. In the biomedical domain, BioNER-LLaMA formulates entity recognition as an instruction-following generation task~\cite{keloth2024advancing}, NERLlama3.1 performs parameter-efficient multi-dataset instruction tuning~\cite{liu2025resource}, and VANER incorporates external biomedical knowledge into natural-language instructions~\cite{bian2024vaner}. These approaches demonstrate the potential of LLMs for generative BioNER across multiple datasets.

However, natural-language instructions and free-form decoding provide limited explicit control over the structured outputs required by BioNER, including entity boundaries, semantic types, and output formats. Recent code-oriented methods address this issue by expressing extraction schemas and annotation guidelines through structured code. KnowCoder encodes extraction schemas using code representations~\cite{li2024knowcoder}, whereas GOLLIE incorporates annotation guidelines into code-style prompts to improve generalization~\cite{sainz2024gollie}. Nevertheless, these approaches typically adopt a single code formulation, leaving the structural diversity across programming languages underexplored. In contrast, MITE reformulates BioNER as a structure-to-structure generation task and transforms each instance into Python, C++, and Java representations. These representations provide structurally diverse supervision without requiring additional biomedical knowledge and enable ensemble inference across code formats for more robust entity extraction.

\begin{figure*}[t]
    \centering
    \includegraphics[width=1\textwidth]{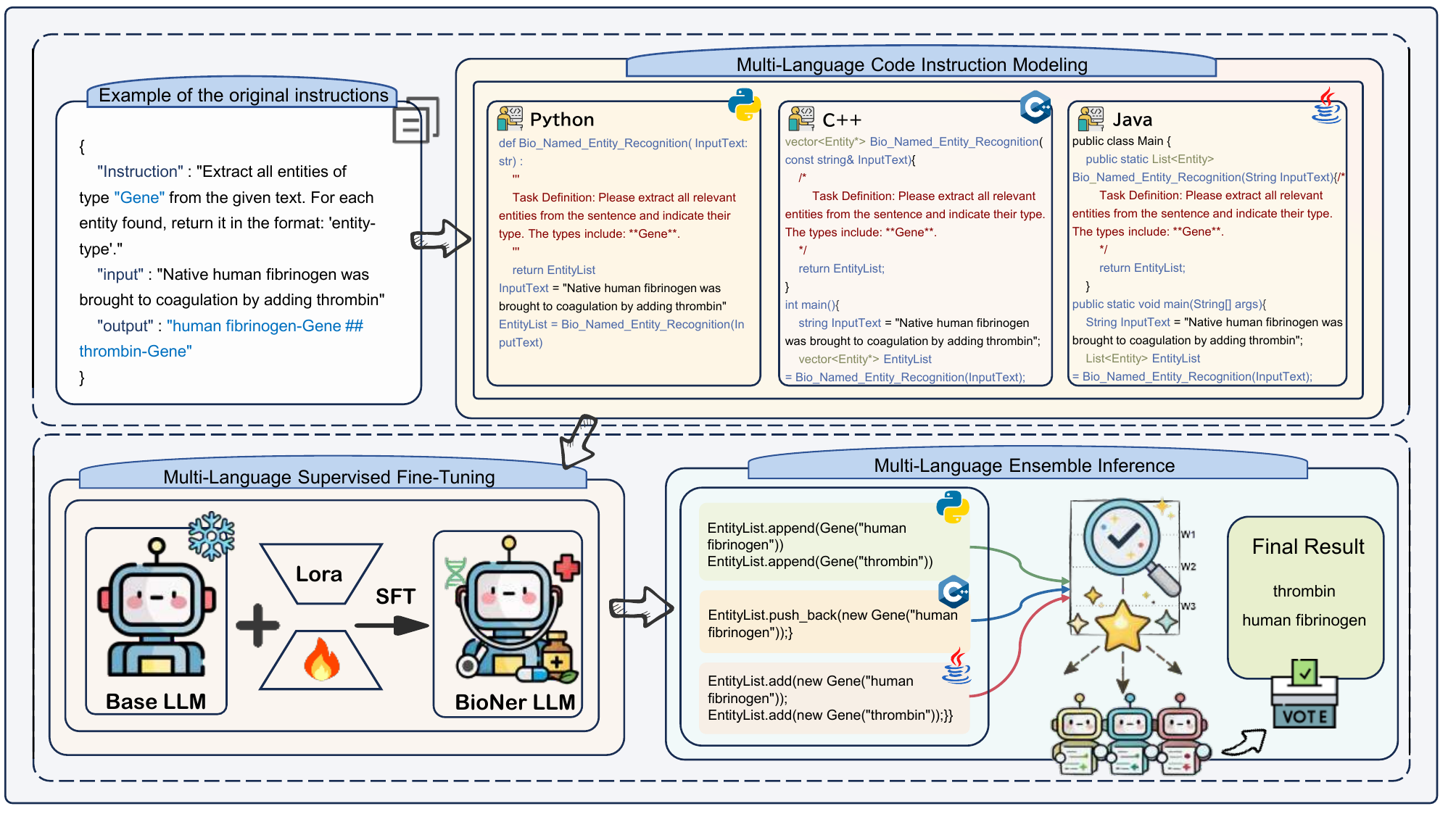}
    \caption{Overview of the MITE framework. MITE converts each BioNER instance into code-formatted instructions in multiple programming languages, fine-tunes an LLM on the resulting multi-language training corpus, and aggregates language-specific predictions through entity-level voting to obtain the final BioNER prediction.}
    \label{fig:model}
\end{figure*}

\section{Methods}

We propose MITE, a Multiple Programming Languages Instruction Tuning and Ensemble method for BioNER. 
MITE reformulates BioNER as a structure-to-structure generation task using code-like representations for both instructions and entity outputs. 
For each instance, MITE constructs Python, C++, and Java representations, fine-tunes an LLM on the resulting multi-language corpus, and ensembles predictions across code formats. 
As shown in Fig.~\ref{fig:model}, MITE consists of four components: 
(1) task formulation, 
(2) multi-language code instruction modeling, 
(3) multi-language supervised fine-tuning, and 
(4) multi-language ensemble inference.

\subsection{Task Formulation}

Given a biomedical token sequence $X=(x_1,x_2,\ldots,x_n)$ and a predefined entity type set $\mathcal{T}$, BioNER aims to extract a set of biomedical entities
$\mathcal{E}=\{(e_j,t_j)\}_{j=1}^{m}$.
Each entity mention $e_j=(x_{s_j},\ldots,x_{k_j})$ is a contiguous text span in $X$, and $t_j \in \mathcal{T}$ denotes its semantic type.
Different from sequence-labeling methods that predict token-level tags, MITE formulates BioNER as a generative structured prediction task, where the model directly produces entity objects specifying entity mentions and their semantic types.

\subsection{Multi-Language Code Instruction Modeling}

Natural-language instructions are easy to construct but often specify the output schema only implicitly, which may lead to unstable entity formats or invalid type assignments. 
To impose stronger structural constraints, MITE represents BioNER instances as code-formatted instructions and outputs. 
The key idea is to express the extraction schema through function signatures, task descriptions, and entity construction statements, so that the model learns to generate structured entity objects rather than free-form textual descriptions.

\subsubsection{Code-Based Instruction}

For each input sentence, we construct a function-based instruction template. 
The function name indicates the BioNER task, the input argument stores the biomedical sentence, and the docstring specifies the target entity types. 
The expected output is represented as entity construction statements added to an \texttt{EntityList}. 
For example, an extracted gene mention is represented as an entity object such as \texttt{Gene("...")}, rather than as an unconstrained text span. 
This design makes the output schema explicit and provides a unified interface for representing entity mentions and semantic types.

\subsubsection{Multi-Language Structural Transformation}
\label{sec:mpl_data_aug}

To further increase structural diversity without introducing additional annotations or external biomedical knowledge, each original BioNER instance is transformed into multiple programming language representations.
Given a BioNER training dataset $\mathcal{D}$, we define:
\begin{equation}
    \mathcal{D}
    =
    \{(X_i,Y_i)\}_{i=1}^{N},
    \label{eq:original_dataset}
\end{equation}
where $X_i$ is the input biomedical sentence, $Y_i$ is the corresponding set of annotated entities, and $N$ is the number of training samples.

For each programming language $l \in \mathcal{L}$, where
$\mathcal{L}=\{\mathrm{Python}, \mathrm{C{+}{+}}, \mathrm{Java}\}$,
we define a language-specific transformation function $\phi_l(\cdot)$:
\begin{equation}
    (C_i^{(l)}, Z_i^{(l)})
    =
    \phi_l(X_i,Y_i),
    \quad l \in \mathcal{L}.
    \label{eq:code_transform}
\end{equation}
Here, $C_i^{(l)}$ denotes the code-formatted instruction in language $l$, and $Z_i^{(l)}$ denotes the corresponding gold structured output in the same language.
Although different languages have different syntactic forms, all transformed outputs preserve the same underlying span-type annotations.

The resulting multi-language training corpus is defined as:
\begin{equation}
    \mathcal{D}_{\mathrm{mpl}}
    =
    \bigcup_{l \in \mathcal{L}}
    \{(C_i^{(l)}, Z_i^{(l)})\}_{i=1}^{N}.
    \label{eq:mpl_dataset}
\end{equation}
This corpus expands each BioNER instance into structurally diverse code representations while keeping the original label semantics unchanged.

\subsection{Multi-Language Supervised Fine-Tuning}

Given the transformed corpus $\mathcal{D}_{\mathrm{mpl}}$, we fine-tune the LLM to generate structured entity outputs conditioned on code-formatted instructions.
Each training sample consists of a language-specific instruction $C_i^{(l)}$ and a gold target output $Z_i^{(l)}$ expressed in the same programming language.
The model is trained autoregressively, where the instruction tokens serve as conditioning context and the loss is computed only over the target output tokens.

The supervised fine-tuning objective is:
\begin{equation}
    \mathcal{L}_{\mathrm{SFT}}
    =
    - \sum_{i=1}^{N}
    \sum_{l \in \mathcal{L}}
    \log P_{\theta}
    \left(
    Z_i^{(l)}
    \mid
    C_i^{(l)}
    \right).
    \label{eq:sft-loss}
\end{equation}
where $P_{\theta}(Z_i^{(l)} \mid C_i^{(l)})$ denotes the autoregressive likelihood of generating the language-specific structured output $Z_i^{(l)}$ conditioned on the corresponding code-formatted instruction $C_i^{(l)}$.
This likelihood is factorized over the target output tokens, while the instruction tokens are used only as conditioning context and are not included in the loss computation.
By jointly optimizing samples from different programming languages, the model learns unified BioNER behavior across heterogeneous code structures.

\subsection{Multi-Language Ensemble Inference}

At inference time, each input sentence is converted into multiple language-specific code instructions. 
The fine-tuned model then generates one structured prediction for each programming language. 
Since different code formats may produce complementary predictions, MITE aggregates them through entity-level majority voting.

Formally, for an input sentence $X_i$, the model produces a prediction $\hat{Y}_i^{(l)}$ for each language $l \in \mathcal{L}$. 
Each prediction is parsed into a set of span-type pairs:
\begin{equation}
    \hat{\mathcal{E}}_i^{(l)}
    =
    \mathrm{Parse}^{(l)}(\hat{Y}_i^{(l)}).
\end{equation}
Two predictions are considered to represent the same entity only when both their mention spans and entity types match.

The final entity set is obtained by majority voting:
\begin{equation}
    \hat{\mathcal{E}}_i^{\mathrm{final}}
    =
    \left\{
    (e,t)
    \mid
    \sum_{l \in \mathcal{L}}
    \mathbb{I}\big[(e,t) \in \hat{\mathcal{E}}_i^{(l)}\big]
    \geq
    \tau
    \right\},
    \label{eq:ensemble_strategy}
\end{equation}
where $\tau=\lceil |\mathcal{L}|/2 \rceil$ is the majority threshold. 
In our setting with three programming languages, an entity is retained if it is predicted by at least two language-specific representations. 
This entity-level voting strategy retains predictions supported by multiple code formats and reduces sensitivity to language-specific prediction disagreements.

\section{Experiments}

\subsection{BioNER Datasets}

We conduct experiments on three widely used BioNER datasets: BC2GM \cite{smith2008overview}, BC5CDR-Chemical \cite{li2016biocreative}, and NCBI Disease \cite{dougan2014ncbi}, covering gene, chemical, and disease recognition, respectively.
Each dataset is independently used for supervised instruction tuning to train a dataset-specific MITE model.
Dataset statistics and splits are reported in Table~\ref{tab:dataset_statistics}.

\begin{table}[ht]      
    \centering
    \caption{Statistics of the BioNER datasets used in our experiments.}
    \label{tab:dataset_statistics}
    \begin{tabular}{l c c c}   
        \toprule
        \textbf{Dataset Name} & \textbf{Train / Dev / Test} & \textbf{Type} & \textbf{Entities} \\
        \midrule
        BC5CDR-Chemical & 4559 / 4580 / 4796 & Chemical & 15828 \\

        BC2GM           & 12632 / 2531 / 5065 & Gene & 24453 \\
        
        NCBI            & 5432 / 923 / 942 & Disease & 6861 \\
        
        \bottomrule
    \end{tabular}
\end{table}

\begin{table*}[t]
    \centering
    \caption{Main experimental results on the NCBI-Disease,
    BC2GM-Gene, and BC5CDR-Chemical datasets, evaluated
    using Micro-F1. The best and second-best results on each
    dataset are highlighted in \textbf{bold} and
    \underline{underlined}, respectively.}
    \label{tab:comprehensive_results_final}

    \begin{tabular*}{\textwidth}{@{\extracolsep{\fill}} l l c c c @{}}
        \toprule
        \textbf{Method} & \textbf{Backbone}
        & \textbf{NCBI (Disease)}
        & \textbf{BC2GM (Gene)}
        & \textbf{BC5CDR (Chemical)} \\
        \midrule
        \multicolumn{5}{c}{\textit{BERT-based Methods}} \\
        \midrule
        BioBERT       & BioBERT     & 88.12 & 81.94 & 83.83 \\
        PubMedBERT    & PubMedBERT  & 86.92 & 83.36 & \underline{93.20} \\
        AIONER        & BioBERT     & \textbf{89.59} & -- & 92.84 \\
        \midrule
        \multicolumn{5}{c}{\textit{LLM-based Methods}} \\
        \midrule
        UniversalNER  & LLaMA2      & 86.96 & 82.42 & -- \\
        InstructUIE   & T5          & 86.21 & 80.69 & -- \\
        BioNER-LLaMA  & LLaMA2      & 88.00 & 83.40 & 92.80 \\
        LLM-MNER      & LLaMA3.1
                      & 88.22 & \underline{83.86} & 92.06 \\
        \textbf{MITE} & LLaMA3.1
                      & \underline{88.33} & \textbf{84.57} & \textbf{93.26} \\
        \bottomrule
    \end{tabular*}
\end{table*}

\subsection{Baselines}

To evaluate the effectiveness of MITE, we compare it against two categories of baseline approaches: BERT-based methods and LLM-based methods. 

\subsubsection{BERT-based Methods}
\begin{itemize}
    \item \textbf{BioBERT} \cite{lee2020biobert} adapts BERT to the biomedical domain through large-scale pretraining on PubMed and PMC.
    \item \textbf{PubMedBERT} \cite{gu2021domain} trains from scratch on PubMed with a biomedical-specific vocabulary.
    \item \textbf{AIONER} \cite{luo2023aioner} integrates annotation labels from multiple datasets to construct a unified sequence labeling task, enhancing multi-task learning performance.
\end{itemize}

\subsubsection{LLM-based Methods}
\begin{itemize}
    \item \textbf{InstructUIE} \cite{wang2023instructuie} uses Flan-T5 to unify information extraction tasks in a text-to-text framework, leveraging multi-task instruction tuning to enhance knowledge sharing and generalization across different extraction tasks.
    \item \textbf{UniversalNER} \cite{zhou2024universalner} distills ChatGPT into LLaMA-2 for NER as a natural language generation task across multiple domains via instruction tuning, using targeted distillation to acquire diverse entity recognition capabilities from teacher-generated annotations.
    \item \textbf{BioNER-LLaMA} \cite{keloth2024advancing} transforms biomedical NER into a generation task using instruction tuning on LLaMA-2 across multiple datasets.
    \item \textbf{LLM-MNER} \cite{calcina2026extracting} explores zero-shot, few-shot, and parameter-efficient fine-tuning strategies for BioNER using large language models.
\end{itemize}

\subsection{Evaluation Metrics}
We evaluate BioNER performance using entity-level Micro-F1 based on exact matching of predicted entity mentions and their types. A prediction is considered correct when both the entity text and its assigned type match the ground-truth annotation.

\subsection{Experiment Settings}
In our experiments, we adopt LLaMA3.1-8B-Instruct as the backbone model for supervised fine-tuning. To achieve parameter-efficient training, we employ LoRA with a rank of 32, an alpha of 64, and a dropout rate of 0.05. Each dataset is independently fine-tuned for 15 epochs using the AdamW optimizer, with a batch size of 10 and a maximum sequence length of 2048 tokens. The learning rate is initialized at 6e-5 and adjusted using a cosine scheduler with a warm-up proportion of 0.05. All experiments are conducted on a single NVIDIA RTX PRO 6000 GPU.

\section{Results and Analysis}

\subsection{Main Results}
Table~\ref{tab:comprehensive_results_final} presents the results on three widely used BioNER benchmarks. Overall, MITE achieves the best performance on BC2GM-Gene and BC5CDR-Chemical, while ranking second on NCBI-Disease, demonstrating competitive performance against both BERT-based and LLM-based approaches. Compared with BERT-based methods pretrained on biomedical corpora, MITE benefits from the strong language understanding and structured generation capabilities of its LLM backbone. It also consistently outperforms representative LLM-based methods, including UniversalNER, BioNER-LLaMA, and LLM-MNER. Unlike these methods that primarily formulate BioNER as natural-language-to-text generation, MITE adopts a code-based instruction paradigm with explicit structural constraints, leveraging code-related structural generation capabilities to facilitate entity boundary recognition and type prediction. These improvements can be attributed to the proposed multiple programming-language instruction tuning and ensemble inference strategies, which enhance prediction robustness and stability.

\begin{table}[t]
    \centering
    \caption{Ablation of individual programming-language representations. Values in parentheses denote differences from the full MITE model.}
    \label{tab:ablation}
    \setlength{\tabcolsep}{15pt}
    \begin{tabular}{l c}
        \toprule
        \textbf{Training Setting} & \textbf{Avg. Score} \\
        \midrule
        MITE$_{\text{w/o Python}}$ & 88.42 \scriptsize{(-0.30)} \\
        MITE$_{\text{w/o C++}}$    & 88.31 \scriptsize{(-0.41)} \\
        MITE$_{\text{w/o Java}}$   & 88.56 \scriptsize{(-0.16)} \\
        \midrule
        \textbf{MITE}              & \textbf{88.72} \\
        \bottomrule
    \end{tabular}
\end{table}

\subsection{Ablation Study}
To evaluate the contribution of each programming language, we remove one language at a time from the training corpus while keeping other settings unchanged. As shown in Table~\ref{tab:ablation}, removing Python, C++, and Java reduces average Micro-F1 by 0.30, 0.41, and 0.16 points, respectively. These consistent performance drops indicate that different languages provide complementary structural information, with gains arising from their joint use rather than a single dominant language. To distinguish structural diversity from data scale, we conduct controlled experiments using the settings in Table~\ref{tab:training_settings}. Single-language $1\times$ settings use one language representation, while their $3\times$ variants expand the corpus through different output orders and docstring descriptions without changing the underlying language structure. The results show that increasing single-language data improves performance by 0.36, 0.35, and 0.21 points for Python, C++, and Java, respectively. In contrast, MITE$_{\text{sampled-train}}$, which maintains the $1\times$ data scale but samples templates across languages, achieves 88.32, outperforming all single-language $1\times$ variants. Moreover, the full MITE achieves the highest score of 88.72, surpassing all single-language $3\times$ settings. These findings suggest that combining multiple programming languages provides benefits beyond simply expanding the training data within a single language.

\begin{table}[t]
    \centering
    \caption{Effect of data scale and programming-language diversity on average Micro-F1. Values in parentheses denote differences from the full MITE model.}
    \label{tab:training_settings}
    \setlength{\tabcolsep}{15pt}
    \begin{tabular}{l c}
        \toprule
        \textbf{Training Setting} & \textbf{Avg. Score} \\
        \midrule
        Python$_{1\times}$                  & 88.11 \scriptsize{(-0.61)} \\
        C++$_{1\times}$                     & 88.03 \scriptsize{(-0.69)} \\
        Java$_{1\times}$                    & 88.21 \scriptsize{(-0.51)} \\
        MITE$_{\text{sampled-train}}$        & 88.32 \scriptsize{(-0.40)} \\
        \midrule
        Python$_{3\times}$                  & 88.47 \scriptsize{(-0.25)} \\
        C++$_{3\times}$                     & 88.38 \scriptsize{(-0.34)} \\
        Java$_{3\times}$                    & 88.42 \scriptsize{(-0.30)} \\
        \midrule
        \textbf{MITE}                       & \textbf{88.72} \\
        \bottomrule
    \end{tabular}
\end{table}

\section{Conclusion}

In this paper, we propose MITE, a Multiple Programming Languages Instruction Tuning and Ensemble method for BioNER. MITE reformulates BioNER as a structure-to-structure generation task by representing both task instructions and entity outputs in code-formatted structures, providing stronger structural constraints for typed entity extraction and better aligning BioNER prediction with the structured patterns learned by LLMs during pretraining. To improve robustness without relying on external biomedical knowledge or additional annotations, MITE transforms each BioNER instance into multiple programming-language representations and aggregates their predictions through entity-level voting. Experiments on three BioNER datasets demonstrate that MITE achieves competitive performance against representative baselines, achieving the best results on two benchmarks, while ablation studies further validate the effectiveness of multi-language representations.

\bibliographystyle{IEEEtran}
\bibliography{citations}

\end{document}